\documentclass[runningheads]{llncs}
\usepackage[T1]{fontenc}
\usepackage{graphicx}
\usepackage{microtype}
\usepackage{algorithmicx,algorithm}
\usepackage{microtype}
\usepackage{xcolor}
\usepackage{adjustbox}
\usepackage{paralist}
\usepackage{multirow,multicol}
\usepackage{tabularx}
\usepackage{colortbl}
\usepackage{tabularx}
\usepackage{algpseudocode}
\usepackage{booktabs}
\usepackage{arydshln}
\usepackage{amsmath}
\usepackage{amssymb}
\begin{document}
\title{Do You Know My Emotion? Emotion-Aware Strategy Recognition towards a Persuasive Dialogue System}
%
%
\author{Wei Peng\inst{1,2}\orcidID{0000-0001-8179-1577} \and
Yue Hu\inst{1,2}\thanks{ Corresponding author.} \and
Luxi Xing\inst{1,2} \and
Yuqiang Xie\inst{1,2} \and
Yajing Sun\inst{1,2}}
\institute{Institute of Information Engineering, Chinese Academy of Sciences, China \and
School of Cyber Security, University of Chinese Academy of Sciences, China}
%
\maketitle              
\begin{abstract}
Persuasive strategy recognition task requires the system to recognize the adopted strategy of the persuader according to the conversation. However, previous methods mainly focus on the contextual information, little is known about incorporating the psychological feedback, i.e. emotion of the persuadee, to predict the strategy. In this paper, we propose a Cross-channel Feedback memOry Network (CFO-Net) to leverage the emotional feedback to iteratively measure the potential benefits of strategies and incorporate them into the contextual-aware dialogue information. Specifically, CFO-Net designs a feedback memory module, including strategy pool and feedback pool, to obtain emotion-aware strategy representation. The strategy pool aims to store historical strategies and the feedback pool is to obtain updated strategy weight based on feedback emotional information. Furthermore, a cross-channel fusion predictor is developed to make a mutual interaction between the emotion-aware strategy representation and the contextual-aware dialogue information for strategy recognition. Experimental results on \textsc{PersuasionForGood} confirm that the proposed model CFO-Net is effective to improve the performance on M-F1 from 61.74 to 65.41.

\keywords{Persuasive dialogue \and Emotional feedback \and Strategy recognition.}
\end{abstract}
\section{Introduction}

Persuasive conversation is an essential area in dialogue systems and has embraced a boom in recent NLP research \cite{DBLP:journals/ijon/ChenGMHP21,Wang2021IncorporatingSK,DBLP:conf/aaai/HideyM18,DBLP:conf/naacl/YangCYJH19}. In a dyadic persuasive dialogue, one party, the persuader, tries to induce another party, the persuadee, to believe something or to do something \cite{DBLP:journals/corr/abs-1912-06745} by a series of persuasion strategies \cite{DBLP:conf/acl/WangSKOYZY19}. However, recognizing the persuasion strategy is challenging in the field of natural language understanding since it needs a deeper understanding of conversation, semantic information, and even the psychological feedback of speakers \cite{DBLP:journals/ijon/ChenGMHP21,DBLP:journals/aai/PrendingerI05,DBLP:journals/corr/abs-2204-12749}. Furthermore, dialogue systems can utilize the predicted historical strategy chains to guide the dialogue generation task.

\begin{figure}[t]
	\centering
	\includegraphics[width=0.5\textwidth]{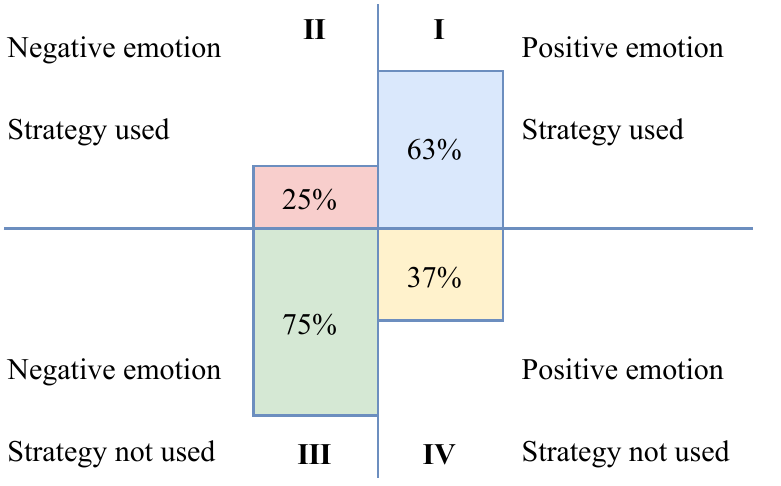}
	\caption{Statistics in the dataset to show relationships between the emotion and strategy. 
	}
	\label{fig:example2}
\end{figure}

To make persuasive strategy prediction, mainstream studies \cite{DBLP:journals/ijon/ChenGMHP21,DBLP:journals/corr/abs-2009-13902} focused on the conversational context to recognize strategies. Some work considered resistance strategies to model the strategy conversations, such as \cite{DBLP:journals/corr/abs-2101-10545} and \cite{DBLP:conf/emnlp/TianSLY20}. However, analyzing and understanding speaker's psychological emotion is an essential job \cite{DBLP:journals/corr/abs-1906-09774,DBLP:journals/iwc/PartalaS04} to fully understand the conversation and help persuader to adopt appropriate strategies. Previous methods do not take the persuadee-aware emotional feedback into account thereby fail to model the emotion-aware human persuasive dialogue systems. To illustrate the importance of emotional feedback, the statistics in the dataset have shown the relationships between the emotion and strategy in Fig. \ref{fig:example2}. The whole plane is divided into four quadrants. As shown in quadrant-I, if the persuadee shows positive emotion after using the strategy $\mathcal{ X }$, the probability of strategy $\mathcal{ X }$ continuing to be used is 63\% in the following conversation. Similarly, in quadrant-III, when the persuadee shows negative emotion, the probability of the strategy not being used in the subsequent conversation is 75\%.

\begin{figure*}[!]
	\centering
	\includegraphics[width=1.0\textwidth]{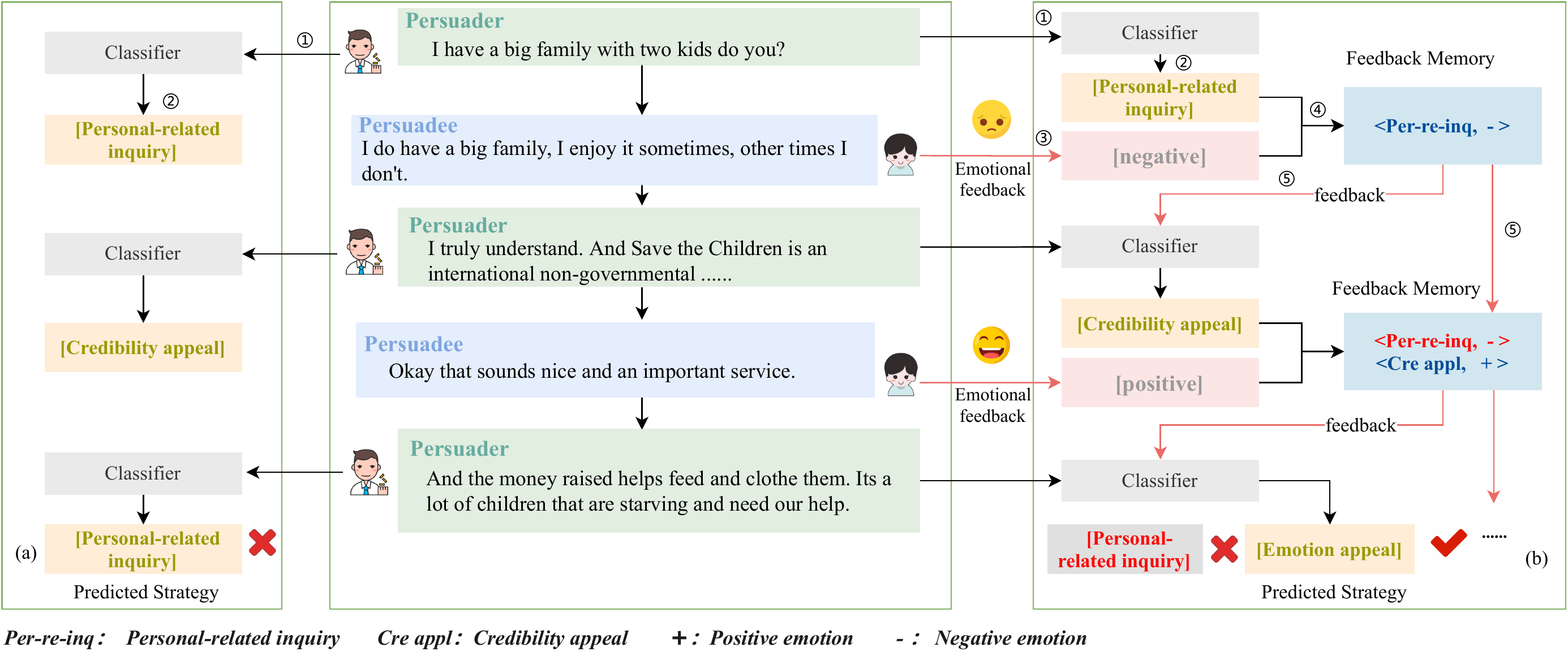}
	\caption{An example is to compare previous work (a) that utilized the contextual information and our work (b) that considers emotional feedback of the persuadee to recognize the strategy. \textcircled{\small{n}} indicates the order of processes.}
	\label{fig:example1}
\end{figure*}

Statistical results indicate that if the strategy obtains positive feedback, it can be given priority in the future. On the contrary, the strategy should be paid less attention \cite{BaronCohen2004TheEQ,2000Understanding}.
To present the discrepancy between the previous work (a) and ours (b), an example is shown in Fig. \ref{fig:example1}. Specifically, in the third turn, (a) outputs the wrong prediction \textit{personal-related inquiry} which has received the negative emotional feedback in the previous turn. Therefore, it would be more appropriate to give priority to a different strategy based on the emotional feedback. This leaves us with: \textit{How to model and incorporate emotional feedback into the contextual dialogue information to achieve a better strategy recognition? }

In this paper, the proposed Cross-channel Feedback memOry Network (CFO-Net) leverages persuadee's emotional feedback to iteratively measure the potential benefit of historical strategies, and further the updated representations of strategies are used to guide the strategy recognition. Specifically, the novel feedback memory module designs strategy pool and feedback pool to process and store the historical strategies and update the strategy weight based on the emotional feedback, respectively. Furthermore, the emotion-aware strategy representation and the contextual information are interacted by the designed cross-channel fusion predictor to make the final strategy recognition.

The contributions can be summarized as follows:
\begin{itemize}
	\item We propose a CFO-Net to leverage persuadee's emotional feedback to measure the potential benefit of historical strategies, and incorporate them into context with cross-channel fusion predictor for persuasive strategy recognition.
	\item A novel feedback memory module is presented to keep track of the historical strategies and further to obtain the emotion-aware strategy representation in a dynamic and iterative manner.
	\item Experiments on the dataset show that the CFO-Net has strong competitiveness with baselines and improves the performance of strategy recognition significantly.
\end{itemize}

\section{Related Work}

\label{rl-class}
\subsection{Non-Collaborative Dialogue}
In collaborative dialogue, systems collaborate and communicate with each other to achieve a common goal \cite{DBLP:conf/acl/HeBEL17}.
A large number of researches \cite{DBLP:journals/corr/abs-1709-05411,larionov2018tartan,Wang2021IncorporatingSK} have shown remarkable advancement in the collaborative setting. However, they are out of scope when applied to non-collaborative settings like negotiation or persuasion. For the negotiation task, two agents have a conflict of interest but must strategically communicate to reach an agreement like a bargaining scenario \cite{DBLP:conf/emnlp/HeCBL18}. In this paper, the main focus is on the persuasive scenario, where the persuader tries to induce people to donate \cite{DBLP:conf/acl/WangSKOYZY19}. The persuasion strategies are identified as ten categories in \cite{DBLP:conf/acl/WangSKOYZY19} that can be divided into two types, 1) persuasive appeal and 2) persuasive inquiry. Specifically, persuasive appeal contains seven strategies (Logical appeal, Emotion appeal, Credibility appeal, Foot-in-the-door, Self-modeling, Personal story and Donation information). For example, personal story refers to the strategy of using
narrative examples to state someone’s donation experiences or the beneficiaries’ positive outcomes, which can encourage others to follow the actions. In addition, the three strategies (Task-related inquiry, Personal-related inquiry and Source-related inquiry) belong to persuasive inquiry, which builds better interpersonal relationships by asking questions. For example, source-related inquiry asks whether the persuadee knows about the organization (i.e., the source in our specific donation task).

\subsection{Persuasive Dialogue Systems}
Persuasive dialogue systems, which have come to increasing attention, aim to change people’s behaviors by persuasive strategies \cite{andre2000automated,DBLP:journals/dsonline/Hilf03,OinasKukkonen2008TowardsDU,DBLP:journals/aiedu/YuanMG08}. For instance, \cite{DBLP:conf/argmining/HideyMHMM17} proposed a two-tiered annotation scheme to distinguish claims in an online persuasive forum. \cite{DBLP:conf/aaai/HideyM18} proposed to predict persuasiveness by modeling argument sequence in social media. \cite{DBLP:conf/naacl/YangCYJH19} designed a hierarchical neural network to identify persuasion strategies. Furthermore, some work focused on the contextual information and modeled the utterances to recognize the strategy. \cite{DBLP:journals/corr/abs-2009-13902} explored and quantified the role of context for different aspects of dialogue for strategy prediction.
\cite{DBLP:journals/ijon/ChenGMHP21} introduced a transformer-based approach coupled with Conditional Random Field for strategy recognition. A few work considered the resistance strategies to model the strategy conversations like \cite{DBLP:conf/emnlp/TianSLY20} and \cite{DBLP:journals/corr/abs-2101-10545}. The Hybrid-RCNN \cite{DBLP:conf/acl/WangSKOYZY19} extracted sentiment embedding features (pos, neg, neu) but did not include the emotion in the history modeling, and ignored the corresponding strategy. 
To overcome these defects, we present the CFO-Net to leverage the emotional feedback to iteratively measure the potential benefits of strategies and incorporate them into the context.

\section{Approach}
\begin{figure*}[htbp]
	\centering
	\includegraphics[width=0.99\textwidth]{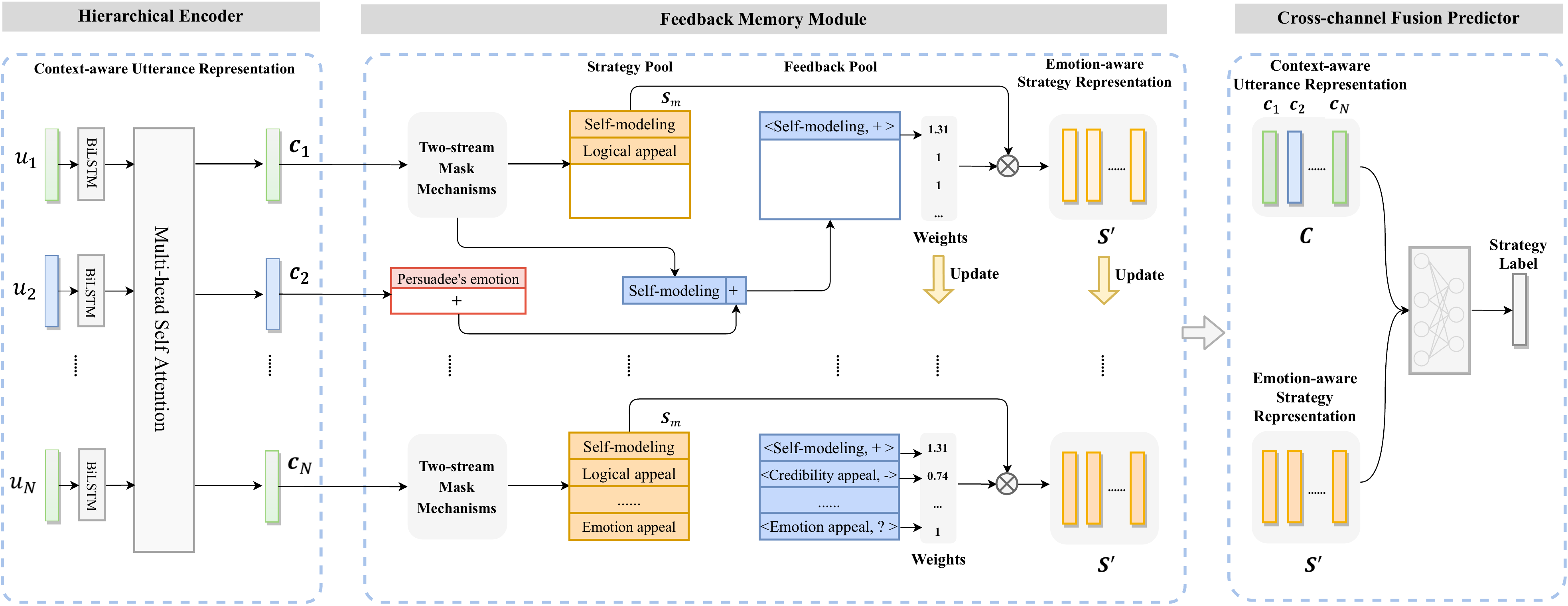}
	\caption{The overview of CFO-Net, which consists of hierarchical encoder, feedback memory module and cross-channel fusion predictor. Green and blue vertical bars mean the utterances of persuader and persuadee. The emotion-aware strategy representation is updated iteratively based on the strategy pool and feedback pool. }
	\label{fig:model}
\end{figure*}

As shown in Fig. \ref{fig:model}, the proposed model consists of three components: \textbf{(a)} \textbf{a hierarchical encoder}, which encodes the contextual dependency with the multi-head attention to capture the semantic information, \textbf{(b)} \textbf{a feedback memory module}, which models the interaction between the strategy pool and the feedback pool to obtain emotion-aware strategy representation, and \textbf{(c)} \textbf{a cross-channel fusion predictor}, which makes an interaction between the emotional feedback and the contextual information, and outputs the final result. Each component is described in the following.

\subsection{Hierarchical Encoder}
The hierarchical encoder uses a \textbf{Bi-directional LSTM} (BiLSTM) \cite{Hochreiter1997LongSM} or \textbf{BERT-style encoder} \cite{DBLP:conf/naacl/DevlinCLT19,DBLP:journals/corr/abs-1907-11692,DBLP:conf/acl/LiCCZ0ZL22}, which capture the temporal features within the words.
Then, the \textbf{Multi-head Attention} aims to explore the semantic information at different granularity.

\subsubsection{Utterance Encoder with BiLSTM} 
The Utterance Encoder vectorizes an input utterance. Given a historical conversation $C = (u_1, u_2, \dots, u_{N})$ a set of ${N}$ utterances, where $u_i = (x_{i,1}, x_{i,2}, \dots, x_{i,T})$ that consists of a sequence of $T$ words, $u_{N}$ indicates the utterance of the persuader, which is uesd to predict the persuasion strategy. A BiLSTM is utilized to encode each word $x_{i,t}$ in the utterance $u_i \in C$, leading to a series of context-aware hidden states $(\textbf{h}_{i,1}, \textbf{h}_{i,2}, \dots, \textbf{h}_{i,T})$, $\textbf{h}_{i,t} = {\rm{concat}}[~\overrightarrow{\textbf{h}_{i,t}}~;~ \overleftarrow{\textbf{h}_{i,t}}~]$.

The last hidden state $\textbf{h}_{i,T}$ is considered to get the utterance-level representation. (Note: the representation of the [CLS] is used as the utterance-level representation in BERT-style encoders). Therefore, the set of $N$ utterances in $C$ can be represented as $\textbf{H} = (\textbf{h}_{1,T}, \textbf{h}_{2,T}, \dots, \textbf{h}_{{N},T})$.

\subsubsection{Utterance-level Multi-head Attention}
To explore the semantic information at different granularity, the multi-head attention \cite{DBLP:conf/nips/VaswaniSPUJGKP17} is adopted as shown in Eq. (\ref{equ:transformer}). $\textbf{c}_{i}$ indicates the representation of $i$-th utterance:
\begin{equation}\label{equ:transformer}
\textbf{c}_{i} = {\rm{Multi\verb|-|head~Attention}}(\textbf{h}_{i,T})
\end{equation}

\subsection{Feedback Memory Module}
The proposed feedback memory module is composed of three novel factors. \textbf{(i) Strategy Embedding} represents the features of strategies which will be continuously updated to capture persuasive strategy features. \textbf{(ii) Strategy Pool} temporarily processes and stores all the possible historical strategies for future reference. \textbf{(iii) Feedback Pool} considers the emotional feedback of the persuadee to measure the potential benefits of strategies and updates the strategy weight $\gamma$.
Finally, the strategy pool and feedback pool are interacted to obtain the emotion-aware strategy representation for later strategy recognition.
\subsubsection{Strategy Embedding}
In the feedback memory module, a randomly initialized strategy embedding is defined to represent the strategy features as $\textbf{S} \in \mathbb{R}^{L \times d}$, where $L$ is the number of the strategy labels and $d$ indicates the dimension. The strategy embedding will be continuously updated to capture persuasive strategy features. Specificly, CFO-Net selects the appropriate strategies (i.e. top-$k$) based on the context from strategy embedding and stores them into the strategy pool with the context-aware softmax function that shown in Eq. (\ref{Eq:linear}). 

\begin{figure}[t]
	\centering
	\includegraphics[width=0.6\textwidth]{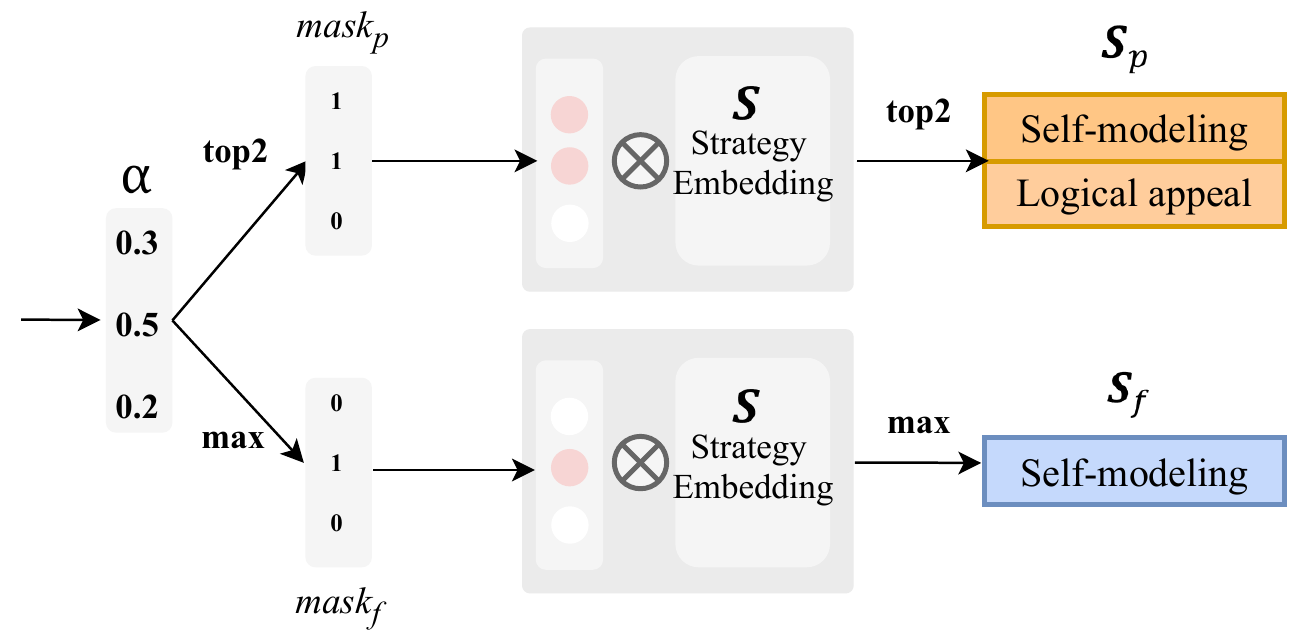}
	\caption{The two-stream mask mechanisms are defined in the feedback memory module.}
	\label{fig:push}
\end{figure}
\subsubsection{Strategy Pool}
Strategy pool aims to process and store the possible historical strategies for future reference. 
As shown in Fig. \ref{fig:push}, to achieve the selection of strategies and prevent gradient truncation, \textbf{two-stream mask} mechanisms are defined in the following:
\begin{itemize}
	\item $mask_p$: The selected strategies (i.e. top-$k$) are stored into the strategy pool to reserve the possible historical strategies (here, size is set to $10$).
	\item $mask_f$: The best strategy of the current moment is stored into the feedback pool. 
\end{itemize}

Specificlly, the module first outputs a probability distribution $\alpha$ of the strategy based on the contextual information, as:
\begin{equation}\label{Eq:linear}
\alpha = {\rm softmax}({\rm MLP}([\textbf{c}_1; \dots; \textbf{c}_{N}]))
\end{equation}

Then, the $mask_p$ is obtained based on the $\alpha$ with the top-$k$ operation where $k$ is a hyper-parameter, and $mask_{f}$ is obtained when $k=1$. The strategies $\textbf{S}_p$ which contain multiple possible strategies are stored into the strategy pool, as:
\begin{equation}
\textbf{S}_p = \textbf{S} \odot (mask_p\otimes \mathbf{e}_d)
\end{equation}
where $\odot$ is element-wise multiplication, $(\cdot \otimes \mathbf{e}_d)$ produces a matrix by repeating the vector on the left for $d$ times \cite{DBLP:conf/iclr/Wang017a}.

The strategies ${\textbf{S}}_m$ in the strategy pool are obtained by making a concatenation with the stored strategies $\textbf{S}_p$. Similarly, the  strategy $\textbf{S}_f$ that stored into the feedback pool is formulated as:
\begin{equation}
\textbf{S}_f = \textbf{S} \odot (mask_{f}\otimes \mathbf{e}_d)
\end{equation}

\subsubsection{Feedback Pool} 
The purpose of the feedback pool is to update the strategy weight $\gamma$ dynamically to
record the emotional feedback of the persuadee towards the strategy. The tuple \{\textit{strategy, emotion}\} stored in the pool calculates the strategy weight $\gamma \in \mathbb{R}^{L}$ that is used to obtain the subsequent emotion-aware strategy representation. Firstly, the representation of utterance $\textbf{c}_i$ is considered to predict the emotional label $ \textbf{y}^e \in \{pos, neu, neg\}$ of the persuadee, as:
\begin{equation}\label{Eq:emotion}
\textbf{y}^e = {\rm softmax}({\rm MLP}([\textbf{c}_1; \dots; \textbf{c}_{N-1}]))
\end{equation}
where $\textbf{c}_{N-1}$ indicates the ${(N-1)}^{th}$ utterance spoken by the persuadee.
\begin{figure*}[htbp]
	\centering
	\includegraphics[width=1.0\textwidth]{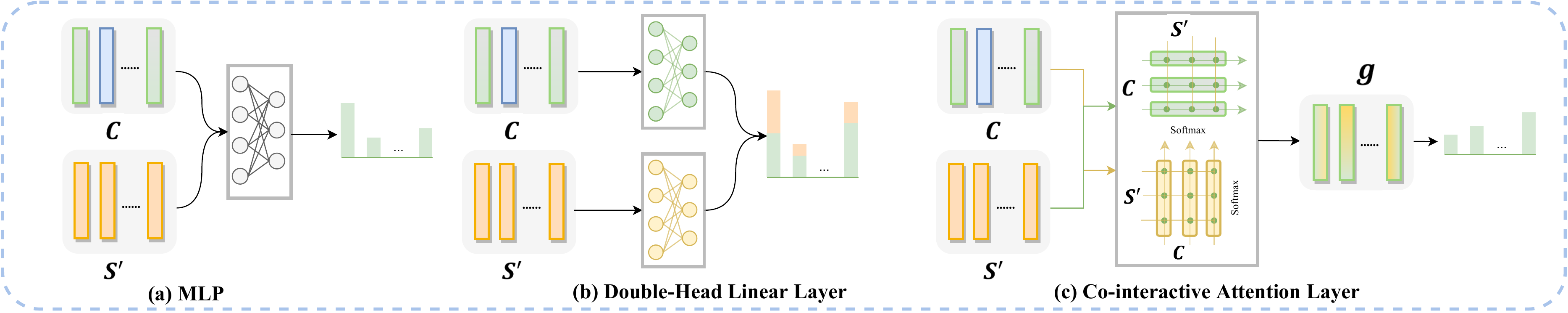}
	\caption{The three different cross-channel fusion mechanisms that include (a) MLP \cite{DBLP:conf/cvpr/NguyenO18}, (b) double-head linear layer and (c) co-interactive attention layer.}
	\label{fig:model-fusion}
\end{figure*}

Then, the weight $\gamma$ is assigned based on the score of the predicted emotion and stream $mask_{f}$. To enhance the generalization of the model, soft weight $\gamma \in \mathbb{R}^{L}$ (randomly initialized with an {all-one vector} at the first) can be defined as: 
\begin{equation}
\label{Eq:gamma}
\gamma_i=
\begin{cases}  \gamma_i + {mask_{f}}\cdot \mu \exp^{-\zeta}  &if \quad {\rm pos};\\
\gamma_i \qquad \qquad \quad \qquad &if \quad {\rm neu};	\\
\gamma_i - {mask_{f}}\cdot \mu \exp^{-\zeta}  &if \quad {\rm neg};
\end{cases}
\end{equation}
where the scalar parameter $\mu$ controls the proportion of $\exp^{-\zeta}$ that guarantees to be greater than zero. For the first condition, the weight of $\gamma$ increases when $\zeta$ becomes smaller. To this end, we intuitively set the confidence factor $\zeta$ that depends on the score of emotion $\textbf{y}^e$, as:
\begin{equation}
\zeta = (1-{y}^e_{x})
\end{equation}
where ${y}^e_{x}$ is a scalar that indicates the score of the $x \in \{pos, neu, neg\}$ emotion. Finally, the emotion-aware strategy representation $\textbf{S}^{\prime}$ is modeled as:
\begin{equation}
\label{Eq:9}
\textbf{S}^{\prime} = \gamma \cdot \textbf{S}_{m}
\end{equation}

\subsection{Cross-channel Fusion Predictor}
In this section, the predictor aims to make a recognition of the strategy. Three main types of fusion mechanisms are designed for horizontal comparison in Fig. \ref{fig:model-fusion}. The mechanisms are introduced to fully interact the psychological feedback with the contextual dialogue information. And the predictor outputs the fusion distribution which captures the profound relationships between two sources.

\subsubsection{Multi-layer Perceptron} 
An MLP can obtain the integrated representation automatically in a simple fashion,
as:
\begin{equation}\label{equ:mlp}
\textbf{g} = {\rm{MLP}}([\textbf{c}_1; \dots; \textbf{c}_{N}; \textbf{s}_1^{\prime}; \dots; \textbf{s}_L^{\prime}])
\end{equation}

The predicted distribution of the strategy $\textbf{y}^s$ can be defined as follows:

\begin{align}\label{equ:soft}
\textbf{y}^s = {\rm{softmax}} &(\textbf{W}^{s}{ \textbf{g} } + \textbf{b}_{s})
\end{align}
where $\textbf{W}^{s} \in \mathbb{R}^{L \times 2d}$ is transformation matrices, $\textbf{b}_{s} \in \mathbb{R}^{L}$ is the bias vector, $L$ is the number of the labels.

\subsubsection{Double-head Linear Layer} 
To achieve the fusion of two probability distribution, a double-head linear layer is designed for prediction. Specifically, we introduce two MLPs to calculate respective probabilities and then combine them, as:
\begin{align}\label{equ:g1}
\textbf{y}^s_1 = {\rm{softmax}} (&{\rm{MLP}}~([\textbf{c}_1; \dots; \textbf{c}_{N}]))	\\
\textbf{y}^s_2 = {\rm{softmax}} (&{\rm{MLP}}~([\textbf{s}_1^{\prime}; \dots; \textbf{s}_L^{\prime}]))	\\
\textbf{y}^s = {\rm{soft}} &{\rm{max}}(~\textbf{y}^s_1 + \textbf{y}^s_2~)
\end{align}
where $\textbf{y}^s_1 \in \mathbb{R}^{L} $ and $\textbf{y}^s_2 \in \mathbb{R}^{L} $, $\textbf{y}^s$ is the final predicted distribution of the strategy.

\subsubsection{Co-interactive Attention Layer} 
Motivated by attention mechanism \cite{DBLP:conf/emnlp/LuongPM15,DBLP:conf/iclr/SeoKFH17,peng2021aper}, the co-interactive attention layer is proposed to effectively model mutually relational dependency.
In this layer, attentions are computed in two directions: from $\textbf{C}=(\textbf{c}_1, \dots, \textbf{c}_{N})$ to $\textbf{S}^{\prime}=(\textbf{s}_1^{\prime}, \dots, \textbf{s}_L^{\prime})$ as well as from $\textbf{S}^{\prime}$ to $\textbf{C}$.

Specifically, the layer first yields a shared similarity matrix $A \in \mathbb{R}^{{N} \times L}$, between $\textbf{C}$ and $\textbf{S}^{\prime}$. $\textbf{A}_{ij}$ indicates the similarity between $i$-th context-aware utterance and $j$-th emotion-aware strategy, as:
\begin{equation}\label{Eq:sim}
{\bf A}_{ij} = \mathcal{F}({\bf C}_{:i}, {\bf S}^{\prime}_{:j})
\end{equation}
where $\mathcal{F}$ is a dot product function, 
${\bf C}_{:i}$ is $i$-th row vector of ${\bf C}$, and
${\bf S}^{\prime}_{:j}$ is $j$-th row vector of ${\bf S}^{\prime}$.

The attention weights and the attended vectors can be obtained in both directions. Firstly, considering the direction from $\textbf{S}^{\prime}$ to $\textbf{C}$, the attention weight is computed by ${\bf a}_i = \mathrm{softmax}({\bf A}_{i:}) \in \mathbb{R}^L$,
and subsequently context-aware utterance vector is $\tilde{{\bf C}}_{:i} = \sum_j {\bf a}_{ij} {\bf S}^{\prime}_{:j}$. Similarly, the attention weight ${\bf b}_j = \mathrm{softmax}({\bf A}_{:j}) \in \mathbb{R}^{N}$,
and updated emotion-aware strategy vector is $\tilde{{\bf S}}^{\prime}_{:j} = \sum_i {\bf b}_{ij} {\bf C}_{i:}$.

Finally, the context-aware utterance representation and emotion-aware strategy representation are combined to yield $\textbf{g}$ and $\textbf{y}^s$ like Eq. (\ref{equ:mlp}) and Eq. (\ref{equ:soft}), as:
\begin{equation}\label{equ:bidaf}
\textbf{y}^s = {\rm{softmax}} (\textbf{W}^{s}{ \textbf{g} } + \textbf{b}_{s})
\end{equation}

\subsection{Training}
The objective of strategy and emotion prediction can be formulated as:
\begin{align}
\mathcal { L } _ { s } = - &\sum _ { i = 1 } ^ {  D} \hat { {\bf{y}} } _ { i } ^ { s } \log \left( {\bf{y}} _ { i} ^ {  s } \right) \\
\mathcal { L } _ { e } = - &\sum _ { i = 1 } ^ {  D}  \hat { {\bf{y}} } _ { i } ^ { e } \log \left( {\bf{y}} _ { i} ^ {  e } \right) 
\end{align}
where $D$ is the number of the training data, ${\hat { {\bf{y}} } _ { i } ^ { s } }$ and $ {\hat { {\bf{y}}} _ { i } ^ { e } }$ are gold strategy label and sentiment label, respectively. The joint objective function $\mathcal{L_\theta}$ is formulated with the hyper-parameters $\beta$ as,  $\mathcal{L_\theta}=\beta_1\mathcal{L}_{s}+\beta_2\mathcal{L}_{e}$.

\section{Experiments}
\subsection{Experimental Setting}
\label{dataset}

\noindent
\textbf{Dataset \& Evaluation Metric} Considering there is no emotional score in other dataset, we focus on the \textsc{PersuasionForGood} \cite{DBLP:conf/acl/WangSKOYZY19} \footnote{The data are available at: \small{~https://gitlab.com/ucdavisnlp/persuasionforgood}} whose sentiment label can be obtained based on the manually annotated score.
The persuader strategies are identified to ten categories (detail in Section \ref{rl-class}) and one none category. As for the evaluation metric, Precision, Recall, and Macro F1 (M-F1) are used for the strategy recognition and emotion prediction as the dataset is highly imbalanced \cite{DBLP:journals/ijon/ChenGMHP21}. 

\noindent
\textbf{Implementation details}
The BERT-style baselines have the same hyper parameters given on the paper \cite{DBLP:conf/naacl/DevlinCLT19,DBLP:journals/corr/abs-1907-11692}. Adam optimizer \cite{DBLP:journals/corr/KingmaB14} is used for training, with a start learning rate from \{${2}${e-}${5}$, $4$e-$5$, $6$e-$5$, $8$e-$5$\} and mini-batch size from \{${32}$, $64$\}. The epoch of training is set from \{$3$, $5$, ${7}$, $9$\}. The scalar parameter $\mu$ is set from \{${0.2}$, $0.5$\}. $k$ is set to ${2}$ based on the parameter analysis. The historical strategies and emotion will be preprocessed to the two pool. To coordinate the joint training of the two training objectives, we set $\beta_1$ = $\beta_2$ = $0.5$. Tesla V-100 GPU and PyTorch \cite{Paszke2017AutomaticDI} are used to implement our experiments. 

\begin{table*}[!]	
	\centering
	\setlength\tabcolsep{11pt}
	\resizebox{1.0\linewidth}{!}{
		\begin{tabular}{l|ccc|ccc}
			\toprule
			\multicolumn{1}{c|}{\multirow{2}{*}} & \multicolumn{3}{c|}{{Strategy Recognition}}       & \multicolumn{3}{c}{{Emotion Prediction}}      \\  
			\multicolumn{1}{c|}{} 	& {P $\uparrow$}	& {R $\uparrow$}	& {M-F1 $\uparrow$} 	& {P $\uparrow$} & {R $\uparrow$}      & {M-F1 $\uparrow$}     \\ \midrule
			{Hybrid RCNN + All features} \cite{DBLP:conf/acl/WangSKOYZY19}	 	& {62.17*}	& {59.80*}  & {58.76*} 	& {-} & {-}  & {-}      \\
			{RoBERTa$_{large}$ LogReg} \cite{DBLP:journals/ijon/ChenGMHP21}	 	& {64.88*}	& {68.32*}  & {63.15*} 	& {-} & {-}  & {-}   \\
			{RoBERTa$_{large}$ cLSTM} \cite{DBLP:journals/ijon/ChenGMHP21}		& /	& /	& 64.10 	& -   & -   & -    \\
			{RoBERTa$_{large}$ DialogueRNN} \cite{DBLP:journals/ijon/ChenGMHP21}		& /	& /	& {64.30} 	& -		& -	& -	 \\ \midrule
			{RoBERTa$_{base}$} \cite{DBLP:journals/corr/abs-1907-11692}  & 59.58   & 64.39  	& 58.35 	& 53.21 	& 72.05	& 60.41	\\	
			\textbf{{CFO-Net$_{base}$}} 	& {63.29}	& {67.74}	& {62.41}	& {53.08} & {75.22} & {61.94} 	 \\	\cdashline{1-7}[1pt/1pt]
			{RoBERTa$_{large}$} \cite{DBLP:journals/corr/abs-1907-11692}		& {62.69}	& {69.91}	& {61.74}	& {55.49} & {71.30} & {62.11}  \\
			\textbf{{CFO-Net$_{large}$}} 	& \textbf{66.81}	& \textbf{72.28}	& \textbf{65.41}	& \textbf{58.11} & \textbf{75.88} & \textbf{63.91} 	 \\	 \bottomrule
	\end{tabular}}
	\caption{\label{tab:main} Experiments on \textsc{PersuasionForGood} for strategy recognition and emotion prediction. - indicates the baselines don't take emotional feedback into account, therefore the results are none. * indicates the experiments are implemented by ourselves.}	
\end{table*}

\subsection{Experimental Results}
\subsubsection{Baselines}
State-of-the-art models are used as baselines to test the performance. Considering the advantages of pre-trained language models (PLMs), we replace the Bi-LSTM with RoBERTa \cite{DBLP:journals/corr/abs-1907-11692} to strengthen the baseline for fair comparison, as with the work \cite{DBLP:journals/ijon/ChenGMHP21}. The base and large PLMs are used in the main experiments for a complete comparison. To increase training speed, the base PLMs are utilized in other experiments. Other baselines are shown in Table \ref{tab:main}, \cite{DBLP:conf/acl/WangSKOYZY19} considered a hybrid RCNN model to extract textual features. \cite{DBLP:journals/ijon/ChenGMHP21} combined the PLMs with some state-of-the-art models to recognize the strategy of the persuader.

\subsubsection{Main Results}
As depicted in Table \ref{tab:main}, compared with state-of-the-art models and RoBERTa, the performance of our CFO-Net (with double-head linear layer) has gained a lot. The CFO-Net achieves 4.12\% gain on Precision, 2.37\% gain on Recall and 3.67\% gain on M-F1 score compared with RoBERTa$_{large}$, which demonstrates that the psychological feedback of the persuadee is beneficial for the strategy recognition. The M-F1 reaches the decent performance with the RoBERTa DialogueRNN where four tasks are jointly trained, which shows that the CFO-Net can achieve better performance with fewer tasks. As for the emotion prediction task, the CFO-Net also improves the performance, which shows that jointly training the tasks can provide benefits and boost each other.
This phenomenon illustrates that the emotional feedback of the persuadee has the potential to help the process of strategy recognition task. Our code will be released in the link. \footnote{The codes are available at: \small{~https://github.com/pengwei-iie/CFONETWORK}}

\subsection{Ablation Study}
To get a better insight into the components of the CFO-Net, the ablation study is performed in the Table \ref{tab:ablation}. The experiments demonstrate that either component is beneficial to the final results. Note that by removing the feedback memory module, configuration (1) reduces to the RB-base model.

\subsubsection{w/o Feedback Memory Module} In this setting, the feedback memory module is abandoned for exploring the effectiveness of the psychological feedback. From the result, the performance has declined significantly in all metrics, which confirms our hypothesis that introducing the emotion of the persuadee to the strategy recognition is important.
\begin{table*}[!]	
	\centering
	\setlength\tabcolsep{11pt}
	
	\resizebox{1.0\linewidth}{!}{
		\begin{tabular}{l|cccc|cccc}
			\toprule
			\multicolumn{1}{c|}{\multirow{2}{*}} & \multicolumn{4}{c|}{{Strategy Recognition}}       & \multicolumn{4}{c}{{Emotion Prediction}}      \\  
			\multicolumn{1}{c|}{} 	& {P $\uparrow$}	& {R $\uparrow$}	& {M-F1 $\uparrow$} & $\Delta_{(M-F1)}$	& {P $\uparrow$} & {R $\uparrow$}     & {M-F1 $\uparrow$}  & $\Delta_{(M-F1)} $  \\ \midrule
			{CFO-Net + RoBERTa$_{base}$} 	& {63.29}	& {67.74}	& {62.41}		 & {-}	& {53.08} & {75.22} & {61.94}    & {-} \\\midrule
			{ (1)~w/o feedback memory module} & 59.58   & 64.39  	& 58.35  & {-4.06}		& 53.21 	& 72.05	& 60.41 & {-1.54}  \\
			{ (2)~w/o multi-task learning}		& {59.04}	& {65.17}	& {58.50}	& {-3.91} & 53.06 & {72.62} 	 & {60.52} & {-1.42} \\
			{ (3)~w/o cross-channel fusion} 	& {62.44}	& {66.38}	& {60.53}	& {-1.88} & {53.12} & {72.97} 	 & {60.84} 	& {-1.10} \\	\bottomrule
	\end{tabular}}
	\caption{\label{tab:ablation} The results of ablation study on model components. }
\end{table*}

\subsubsection{w/o Multi-task Learning} Multi-task learning considers the mutual connection between tasks by sharing latent representations. Here, the emotion prediction task is removed to see the performance of strategy recognition. In Table \ref{tab:ablation}, the multi-task learning that is jointly training ($\mathcal{L}_{s}$ and $\mathcal{L}_{e}$) can provide benefits, which shows that the training objectives are closely related and boost each other.

\subsubsection{w/o Cross-channel Fusion} The cross-channel fusion combines the {persuader-aware contextual dependency} with {persuadee-aware emotional dependency}. In this setting, these representations are concatenated directly to make a prediction. The results indicate the fusion mechanisms make a contribution to the overall performance.

\subsection{Performances on the Fusion Mechanism}

\begin{figure*}[!]
	\centering
	\includegraphics[width=1.0\textwidth]{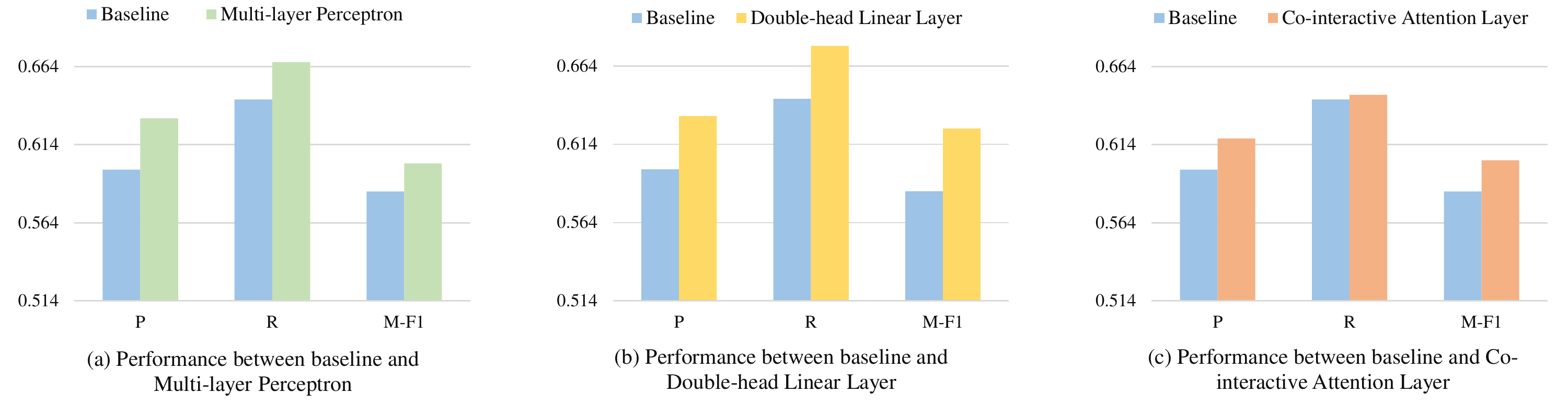}
	\caption{The performances on the fusion mechanism. (a), (b), (c) represent the results between the baseline and the MLP, Double-head Linear Layer and Co-interactive Attention Layer, respectively.}
	\label{fig:fusion}
\end{figure*}

\begin{figure}[!]
	\centering
	\includegraphics[width=0.55\textwidth]{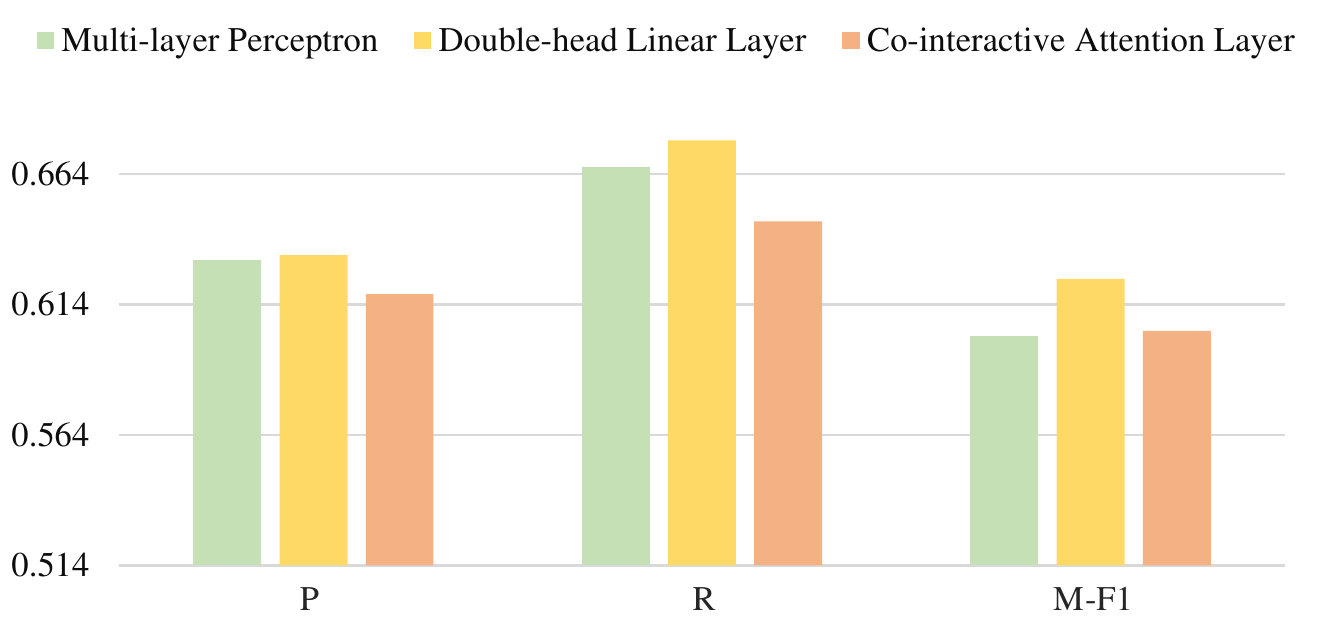}
	\caption{The performances and comparisons on the three different fusion mechanisms.}
	\label{fig:fusion-all}
\end{figure}

The fusion mechanism is adopted to exploit the two types of the interaction, including {persuader-aware contextual dependency} and {persuadee-aware emotional dependency}. To further investigate the effectiveness of these mechanisms, a couple of experiments are conducted from two perspectives, as shown in Fig. \ref{fig:fusion} and Fig \ref{fig:fusion-all}. One is the comparison between three fusion methods and baselines, the other is to consider the horizontal comparison of the fusion mechanisms.

As shown in Fig. \ref{fig:fusion}, the results conclude that the fusion mechanisms incorporating persuadee-aware emotional dependency into persuader-aware contextual dependency can bring consistent improvements and surpass baselines on all evaluation metrics. In addition, Fig. \ref{fig:fusion-all} presents the performances of different fusion mechanisms, in which the double-head linear layer performs best, with the M-F1 score achieving 62.41\%. Surprisingly, the co-interactive attention layer underperforms the double-head linear layer. The phenomenon could be attributed to the fact that the strategy representation and the utterance-level dialogue information belong to different levels of abstract semantic information, leading to the introduction of noise after co-attention operation.

\subsection{Parameter Analysis}
\label{exp:pa}
In the feedback memory module, $k$ is a key hyper-parameter. As shown in Table \ref{table:nosie}, the model will introduce more noise when $k$ is set too large, and the confidence score will become lower, leading to worse performance. On the contrary, the enriched semantic representations of the strategy will be ignored when $k$ is set to one. It shows that although the confidence score is higher, the performance is not the best. The analysis validates that an appropriate $k$ is crucial to the experimental results. 

\begin{table}[!]
	\centering
	\resizebox{0.6\columnwidth}{!}{
		\begin{tabular}{lcccc}
			\toprule
			\textbf{Top-$k$}      & {Top-$1$}		& {Top-$2$}		 & {Top-$3$}	& {Top-$4$} \\ 
			\midrule
			\textbf{M-F1} 		& 60.68		& \textbf{62.41}	& 60.42		& 58.68	\\ 
			\midrule
			\textbf{Confidence Score}      & {\textbf{0.877}}       &{0.473} &{0.326} &{0.242}   \\ 
			\bottomrule
	\end{tabular}}
	\caption{Performance on the hyper-parameter $k$. Confidence score indicates $k^{th}$ average predicted probability.}
	\label{table:nosie}
\end{table}

\subsection{Case Study}
\begin{figure*}[t]
	\centering
	\includegraphics[width=1.0\textwidth]{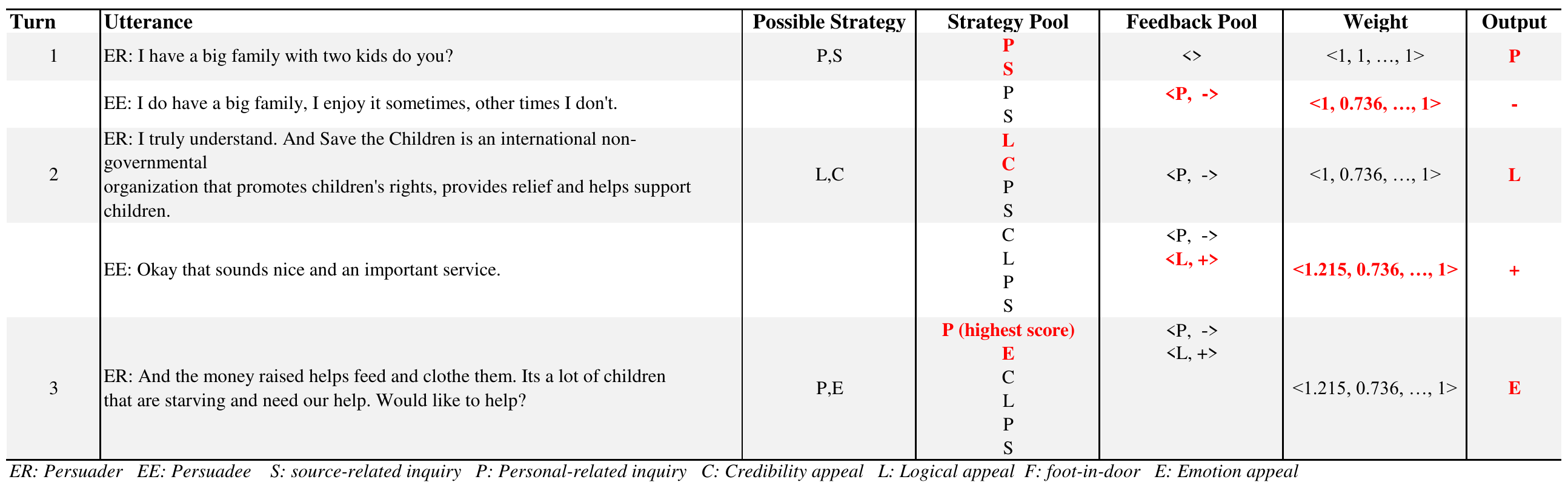}
	\caption{An example to illustrate the process of novel feedback memeory module. The red marker indicates the changes. }
	\label{fig:example}
\end{figure*}

A case study is conducted with the example in Fig. \ref{fig:example} to demonstrate how CFO-Net works when recognizing a strategy. We list the possible strategies, the state of the strategy pool and feedback pool, and the updated weights. In this case, two possible strategies are selected to the strategy pool at a time. Then, the predicted emotion and the strategy with the highest score are stored into the feedback pool in a tuple fashion, such as \textit{$<$Personal-related inquiry, A$>$} where \textit{A} represents \textit{positive} or \textit{neutral} or \textit{negative}. Finally, weights $\gamma$ will be calculated with Eq. (\ref{Eq:gamma}). During the conversation, the strategy recognition not only depends on the contextual dialogue information, but also the emotional feedback of the persuadee. The weights are utilized to compute the emotion-aware strategy representation for the final prediction. In the third turn, the CFO-Net outputs a correct prediction \textit{emotion appeal} rather than \textit{personal-related inquiry} with the highest score calculated by the contextual dialogue information, which indicates that incorporating the emotion-aware strategy representation into the contextual dialogue information is of great importance.

\section{Conclusion}
\label{sec:bibtex}
This paper concentrates on incorporating the psychological feedback (emotion of the persuadee) into the recognition of strategies in the persuasive dialogue. In this paper, we propose a novel Cross-channel Feedback memOry Network (CFO-Net), with a feedback memory module and three different cross-channel fusion mechanisms, to model and explore the historical emotional feedback of persuadee. 
Experimental results and analysis demonstrate that the CFO-Net {has strong competitiveness with baselines and significantly improves the performance of strategy recognition.} For the future work, some other categories of psychological feedback will be considered with BiLSTM-CRF, such as personal character, educational background and so on. These cognitive factors are still worth researching for persuasion dialogue systems. Furthermore, dialogue systems can utilize the predicted historical strategy chains to guide the dialogue generation task.

\section{Acknowledgment}
We thank all anonymous reviewers for their constructive comments and we have made some modifications. This work is supported by the National Natural Science Foundation of China (No.U21B2009).

%
%
%
\bibliographystyle{splncs04}
\bibliography{mybibliography}
%
%
%
%
%
\end{document}